\documentclass{article}

\usepackage[preprint, nonatbib]{ewrl_2023}

\usepackage[utf8]{inputenc} %
\usepackage[T1]{fontenc}    %
\usepackage{hyperref}       %
\usepackage{url}            %
\usepackage{booktabs}       %
\usepackage{amsfonts}       %
\usepackage{nicefrac}       %
\usepackage{microtype}      %
\usepackage{xcolor}         %
\usepackage{graphicx}
\usepackage{latexsym}

\usepackage{subfigure}
\usepackage{caption}
\usepackage{amsmath, amssymb, mathtools,mathrsfs}
\usepackage{wrapfig}
\usepackage{bm}

\newcommand{\smallparagraph}[1]{\smallskip\noindent\textbf{#1}}

\title{Dynamic Size Message Scheduling for Multi-Agent Communication under Limited Bandwidth}

\newcommand{\states}{\ensuremath{\mathcal{S}}}

\newcommand{\jointaction}{\ensuremath{\bm{a}}}
\newcommand{\actions}{\ensuremath{\{\mathcal{A}\}_{i=1}^{n}}}
\newcommand{\actionsindiv}{\ensuremath{\mathcal{A}}}

\newcommand{\discount}{\ensuremath{\gamma}}
\newcommand{\probtransitions}{\ensuremath{\mathbf{P}}} %
\newcommand{\rewards}{\ensuremath{\{\mathcal{R}\}_{i=1}^{n}}}
\newcommand{\rewardsindiv}{\ensuremath{\mathcal{R}}}

\newcommand{\observations}{\ensuremath{\{\Omega\}_{i=1}^n}}
\newcommand{\observationsindiv}{\ensuremath{\Omega}}

\newcommand{\observationfnindiv}{\ensuremath{\mathcal{O}}}
\newcommand{\observationfn}{\ensuremath{\{\mathcal{O}\}_{i=1}^{n}}}
\newcommand{\observation}{\ensuremath{o}}
\newcommand{\jointobs}{\ensuremath{\bm{\observation}}}

\newcommand{\pomgtuple}{\langle \states, \observations, \actions, \probtransitions, \observationfn, \rewards, \allowbreak \discount \rangle}
\newcommand{\fun}[1]{\ensuremath{\mathopen{}\mathclose\bgroup\left(#1\aftergroup\egroup\right)}}

\author{%
  Qingshuang Sun$^{1,2}$ \quad Denis Steckelmacher$^{1}$ \quad Yuan Yao$^{2}$ \quad Ann Nowé$^{1}$ \quad Raphaël Avalos$^{1}$\\
  $^1$ AI Lab, Vrije Universiteit Brussel (Belgium) \\ $^2$ School of Computer Science, Northwestern Polytechnical University (China)  \\
  \texttt{sunqsh@mail.nwpu.edu.cn} \quad \texttt{raphael.avalos@vub.be}
}

\begin{document}

\maketitle

\begin{abstract}
Communication plays a vital role in multi-agent systems, fostering collaboration and coordination. However, in real-world scenarios where communication is bandwidth-limited, existing multi-agent reinforcement learning (MARL) algorithms often provide agents with a binary choice: either transmitting a fixed number of bytes or no information at all. This limitation hinders the ability to effectively utilize the available bandwidth. To overcome this challenge, we present the Dynamic Size Message Scheduling (DSMS) method, which introduces a finer-grained approach to scheduling by considering the actual size of the information to be exchanged. Our contribution lies in adaptively adjusting message sizes using Fourier transform-based compression techniques, enabling agents to tailor their messages to match the allocated bandwidth while striking a balance between information loss and transmission efficiency. Receiving agents can reliably decompress the messages using the inverse Fourier transform. Experimental results demonstrate that DSMS significantly improves performance in multi-agent cooperative tasks by optimizing the utilization of bandwidth and effectively balancing information value.
\end{abstract}

\section{Introduction}

\emph{Multi-agent reinforcement learning} (MARL) has gained significant attention due to its ability to model complex real-world scenarios where multiple autonomous agents interact with the environment. Effective communication among agents plays a crucial role in achieving efficient collaboration, coordination, and the emergence of collective intelligence. Traditional MARL approaches often assume the availability of always-on and unlimited communication channels between agents. However, real-world applications such as multi-robot systems \cite{baldazo2019decentralized, d2020development}, Internet of Things (IoT) networks \cite{kouki2020autonomous}, or distributed sensor networks \cite{zhang2021scalable} often face limited and shared communication bandwidth among multiple agents. Therefore, designing communication protocols that account for the constraints of real-world applications to improve cooperation remains an ongoing challenge.

In deep MARL, communication among agents is commonly and typically generated by a neural network, where the dimension of the communication message is fixed and determined before the training process  \cite{foerster2016learning, sukhbaatar2016learning}. As a result, the size of the communication messages learned by all agents remains constant throughout the task.
While current methods mainly schedule or prune information based on priority to reduce communication load \cite{jiang2018learning, kim2019learning, mao2020learning, pesce2020improving}, this fixed size rigidity presents two notable drawbacks. 
Firstly, it leads to inefficient utilization of available bandwidth when the actual message size is smaller than the output dimension of the neural network. Consequently, valuable bandwidth resources are squandered, resulting in suboptimal communication efficiency. Secondly, and of even greater importance, the fixed size limitation poses a challenge for agents that require transmitting vital messages but are unable to do so due to not being allocated bandwidth. This issue arises particularly when numerous agents concurrently seek to communicate, potentially driven by partial observability. These limitations severely impede the realization of communication's full potential in deep MARL. Therefore, it is imperative to develop flexible and adaptive approaches that can surmount these limitations, enabling efficient and effective communication in multi-agent systems.

\begin{figure*}
    \centering
    \includegraphics[width=.8\linewidth]{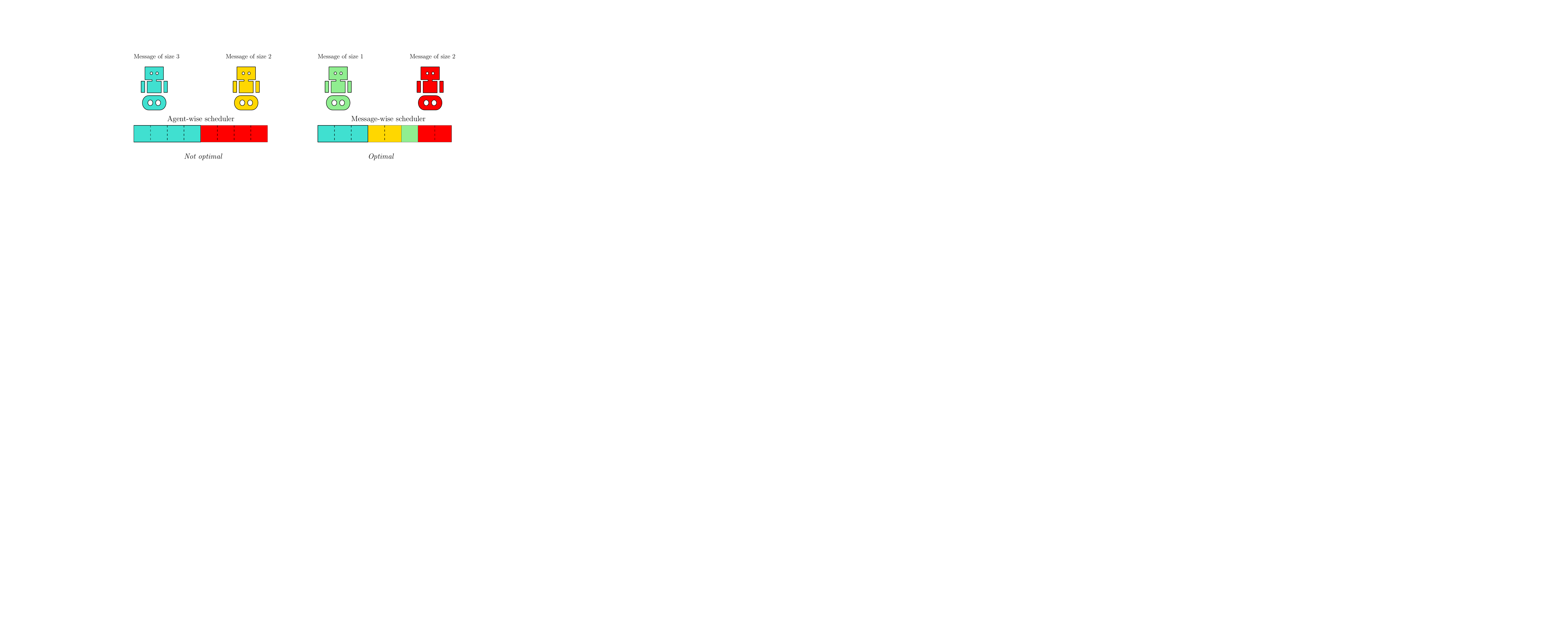}
    \caption{
     Difference between an agent-wise and a message-wise scheduler.  
    }
    \label{fig:scheduler_diff}
\end{figure*}

\begin{figure*}
    \centering
    \includegraphics[width=\linewidth]{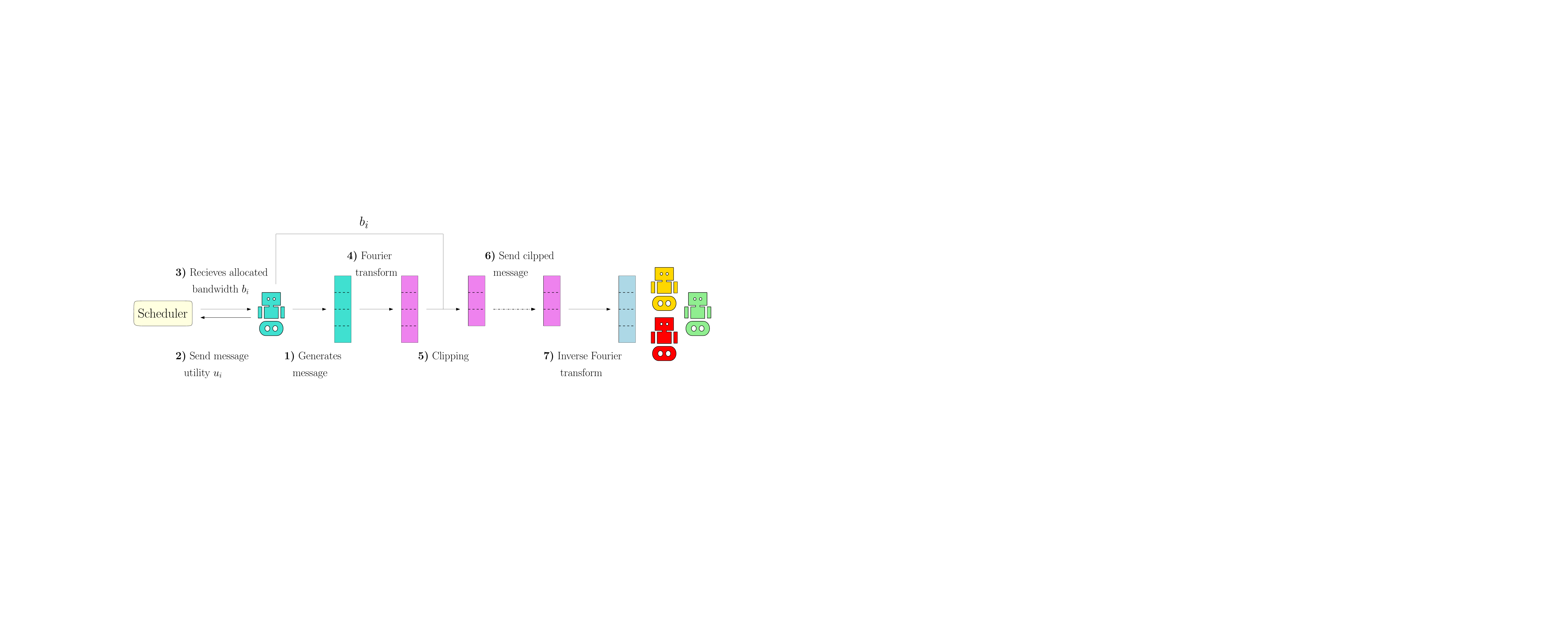}
    \caption{
     Overview of the communication flow.
    }
    \label{fig:com_flow}
\end{figure*}

In this paper, we propose the novel \emph{Dynamic Size Message Scheduling} (DSMS) algorithm. Our approach goes beyond agent-level scheduling and introduces a finer-grained scheduling mechanism that considers the real size of the information to be transmitted. Figure \ref{fig:scheduler_diff} outlines the difference between the general agent-wise scheduling and our message-wise scheduling. DSMS extended granularity allows for improving communication efficiency by exchanging more information under the constraints of the available bandwidth. The core idea of DSMS lies in the fact that the agents first notify the scheduler of the importance of their message, second the scheduler partitions the bandwidth at a byte level, third the agent compresses their original size message to match the allocated bandwidth using a clipped Fourier transformation, finally the receiving agents recover decompress the messages using the inverse Fourier transform. This process reduces the size of messages while maintaining a certain level of fidelity.  Figure~\ref{fig:com_flow} presents an overview of the communication flow.

We evaluate the effectiveness of DSMS by conducting extensive experiments on two partially observable scenarios from multi-agent particle environment (MPE) \cite{lowe2017multi}. Our experimental results demonstrate that DSMS significantly outperforms existing approaches, showcasing its ability to adaptively finer-grained scheduling by allowing agents to send variable-sized messages balancing with the information value to optimize the utilization of limited communication resources.

\smallparagraph{Related work}~
Our work is closely related to prior research in MARL with communication, particularly addressing the challenges posed by limited bandwidth and shared communication mediums. Several existing approaches have focused on enabling agents to learn when to communicate based on the importance of their local observations, such as SchedNet and Gated-ACML \cite{kim2019learning, mao2020learning}. Other methods have explored techniques like temporal locality \cite{zhang2020succinct}, penalty thresholds \cite{hu2021event}, and entropy regularization of messages \cite{wang2020learning} to reduce message transmission. Additionally, shared medium scenarios have been explored in works like MD-MADDPG \cite{pesce2020improving}, which allows agents to access a shared memory space.
However, these existing methods have certain limitations, including fixed message sizes or manual configuration requirements. Furthermore, some approaches rely on uninterrupted access to shared memory, which may not be feasible in practical settings. In contrast, our proposed method overcomes these limitations by introducing a finer-grained scheduling approach that dynamically adjusts message sizes based on their importance, optimizing bandwidth utilization in both limited and shared communication settings.

\section{Background}\label{overview}
\smallparagraph{Partially Observable Markov Games (POMGs)}~ 
\cite{littman1994markov} extend Markov Decision Processes \cite{puterman1990markov} to multiple agents and partial observability. Formally, an $n$-agent POMG is defined as a tuple $\pomgtuple$ with $\states$ the state space, $\observationsindiv_{i}$ the observation space of agent $i$, $\actionsindiv_i$ the action space of agent $i$, $\probtransitions \colon \states \times \Pi_{i=1}^{n} \actionsindiv_{i} \rightarrow \Delta_\states$ the transition probability function, $\observationfnindiv_{i} \colon \states \rightarrow \Delta_{\observationsindiv_i}$ agent $i$'s observation probability function, $\rewardsindiv_{i} \colon \states \times \Pi_{j=1}^{n} \actionsindiv_{j} \rightarrow \mathcal{R}$ agent $i$'s reward function. If all the rewards functions are the same, the environment is fully collaborative and referred to as a Dec-POMDP \cite{oliehoek2016concise, oliehoek2008optimal}. 
Due to the partial observability each agent $i$ needs to condition its policy on its observation-action history $\mu_i \colon \left(\observationsindiv_i, \actionsindiv_i\right)^* \rightarrow \actionsindiv_i$. 
We use LSTMs \cite{gers2000learning} a type of Recurrent Neural Network to represent the history in a fixed-size vector.
In the rest of the paper, we denote with boldface joint elements (i.e. the joint observation $\bm{\observation}$).%

\smallparagraph{Multi-Agent Actor-Critic (MADDPG)}~ 
\cite{lowe2017multi} is a multi-agent Actor-Critic algorithm that adopts centralized training and decentralized execution (CTDE) \cite{oliehoek2008optimal, foerster2018counterfactual, Avalos2022LocalLearningAAMAS} paradigm. It employs a centralized critic to estimate the joint action-value  $Q_{i}^{\mu_i}\left(\jointobs, \jointaction \right)$, which is augmented with the local observation and action of all agents during training. On the other hand, the actors make decisions in a decentralized manner, conditioned only on their local observations.

\smallparagraph{Fourier Transform}~ 
The Fourier transform \cite{boashash1988note, bosi2002introduction} is a fundamental mathematical technique that enables the analysis and transformation of signals between the time domain and the frequency domain. By decomposing a signal into its constituent frequencies, the Fourier transform provides valuable insights into the spectral characteristics of the signal, revealing the magnitudes and phases of its frequency components. This transformation is particularly useful for understanding the frequency content of signals, whether they are periodic or non-periodic. In practical applications, the Fourier transform allows us to identify and extract specific frequencies of interest, enabling tasks such as audio spectrum analysis, image processing, and data compression. It plays a vital role in various fields, including telecommunications, audio and video processing, radar systems, medical imaging, and scientific research. The inverse Fourier transform is used to reconstruct a signal in the time domain from its frequency components, allowing the synthesis of the original signal.

 \begin{figure}
    \centering
    \includegraphics[width=\linewidth]{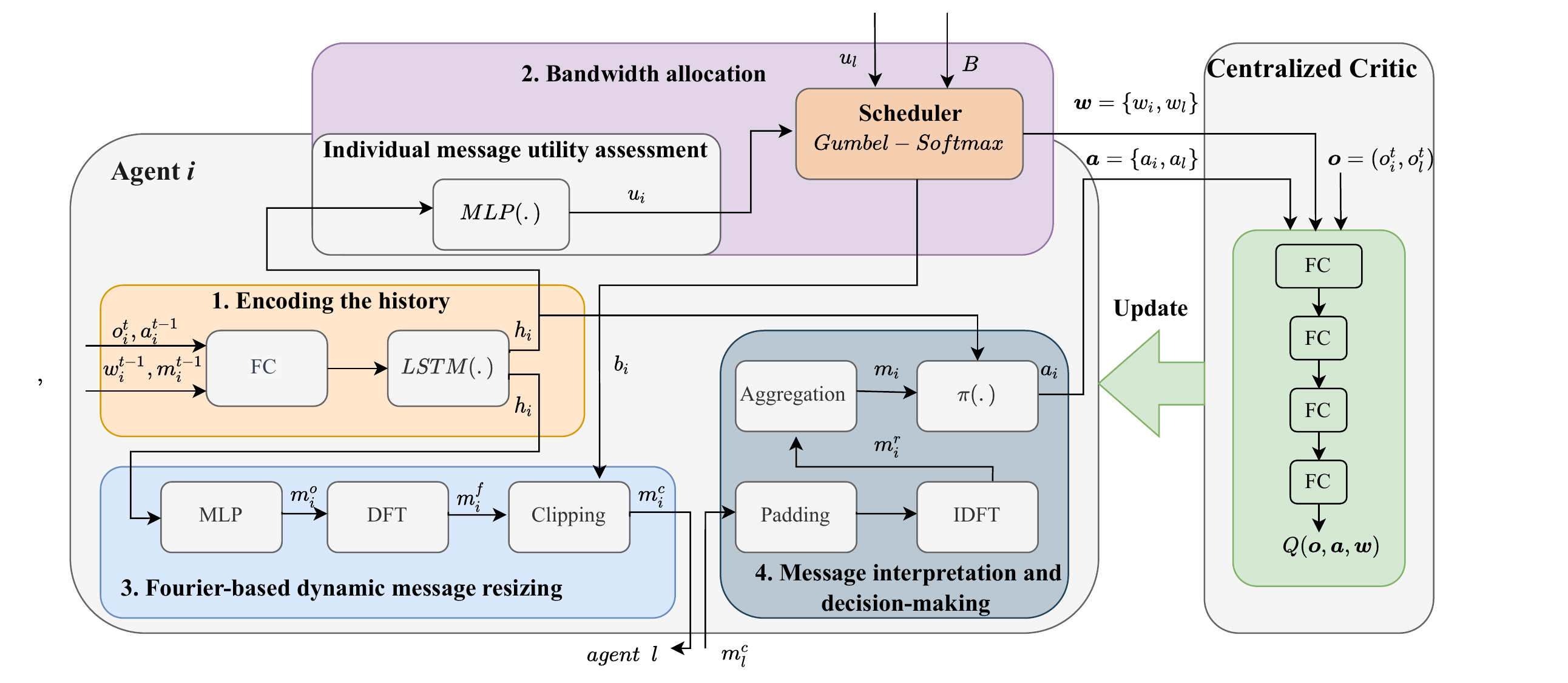}
    \caption{The architecture of DSMS. The illustration of the interaction of two agents $i$ and $l$, where the numbers represent the operational steps of the method. }
    \label{framework}
\end{figure}

\section{Dynamic Size Message Scheduling}\label{decentralized_actor}

In this section, we will discuss the four key components of the proposed algorithm: 1) encoding the history, 2) bandwidth allocation, 3) Fourier-based dynamic message resizing, and 4) message interpretation and decision-making. The architecture of DSMS is illustrated in Figure~\ref{framework}. The training process follows the CTDE paradigm, employing parameter sharing to reduce the number of neural networks that need to be learned and using a centralized critic. To enable agent-specific behavior, the networks are conditioned on an agent ID.

\subsection{Encoding the history}

The first component of DSMS involves utilizing an LSTM to compress the history of each agent into a fixed-size vector. In the context of POMGs, the agent's history typically consists of the previous observation-action pairs. However, in scenarios where agents can communicate, they also need to remember the messages they have received and sent to avoid redundant communication. To address this, we augment the agent's history with two additional elements: the received message at the previous time step, denoted as $m_i^{t-1}$, and the bandwidth weight assigned by the scheduler, represented as $w_i^{t-1}$. By incorporating these elements into the history representation, agents can effectively track the relevant information for communication within the LSTM framework. We denote as $h_i^t$ the output of the LSTM for agent $i$ at time-step $t$.

\subsection{Weight-based bandwidth allocation}
The allocation of bandwidth among agents poses a significant challenge due to the complex nature of assessing the importance of a message, which depends on various parameters that are not directly accessible to the agents. To determine the importance of a message, an agent needs to consider factors such as the information gain compared to previous messages and whether the information is useful to other agents (which is not directly available). Additionally, the agents need to compare the utility of their messages to determine the optimal allocation. To address this challenge, we propose a two-step solution involving individual message utility assessment by each agent followed by centralized scheduling.

\smallparagraph{Individual message utility assessment}~
In the first step, each agent independently assesses the utility of its own message. By leveraging its individual history, which includes the previously allocated bandwidths and the messages received from other agents, an agent can evaluate the potential value of the information it can share with others. To quantify this utility, we employ an MLP that takes the agent's history $h_i$ as input and computes a scalar utility value denoted as $u_i$.

\smallparagraph{Scheduler}~
The scheduler plays a crucial role in allocating the available bandwidth to each agent based on their message utilities. Given the vector of agent message utilities $\bm{u}$, the scheduler employs the Gumbel-Softmax technique \cite{jangcategorical} to transform these utilities into importance weights $\bm{w}$. Although a classic softmax operation could be used for this transformation, we found that the addition of noise in the Gumbel-Softmax improves the training process.

The next step involves converting these importance weights into individual bandwidth sizes $b_i$ for each agent. The equation for computing $b_i$ is as follows:

\begin{align}
    b_i = 2 \times \text{soft\_ceil}\left(\left(\frac{B}{2} - n \right) \times w_i \right) \quad \implies \sum_i b_i \leq B 
\end{align}

In this equation, $B$ represents the total available bandwidth, and $n$ is the number of agents. The $-n$ ensures that there will not be bandwidth overflow from using the ceil operator, and the factor $2$ is required to send complex numbers as explained in the next sub-section. The soft\_ceil operation ensures that gradients can flow during training, and it is defined as:
\begin{align*}
    \text{soft\_ceil}(x) = x + \lceil x \rceil - \text{stop\_grad}(x)
\end{align*}

Importantly, this method guarantees that each agent will have a minimum of two bytes for communication, as the softmax operation ensures that the weights are strictly positive. Although a floor operation could be used instead of ceil, we found out that ceil simplifies the training process.

By applying this procedure, the scheduler effectively assigns bandwidth to each agent based on their message utilities while ensuring that the overall bandwidth constraint is not exceeded.

\subsection{Fourier-based dynamic message resizing}
Based on their history $h_i$, each agent will create a message $m^o_i \in \mathbb{R}^p$ of size $p$ using an MLP. To match the allocated bandwidth $b_i$, agents employ a transformation process that involves Discrete Fourier Transform and clipping.

\smallparagraph{Discrete Fourier Transform (DFT)}~
The motivation behind using Fourier Transform is to map $m^o_i$ into the frequency domain, yielding a condensed representation where the magnitude of the first frequencies contains most of the essential information. The DFT transformation of the original message $m^o_i$ to the frequency message $m^f_i$ is defined as follows, with $m^f_{i,k}$ denoting the $k$-th component of the message.
\begin{align}
    \forall i \in [1, N], \forall k \in [0, p-1], \quad m^f_{i,k} = \sum_{q=0}^{p - 1} m^o_{i,q} \exp{\frac{-2\pi k q j}{p}} \in \mathbb{C}
\end{align}

In this equation, the symbol $j$ represents the imaginary unit, which denotes the square root of -1, and $\mathbb{C}$ refers to the space of complex numbers.

It is important to note that the DFT of a real-valued signal, such as $m^f_i$, exhibits Hermitian symmetry. This symmetry implies that the frequency components at negative frequencies can be derived from their corresponding symmetric components at positive frequencies. As a result, the frequency components of $m^f_{i,k}$ for $k > \lceil\frac{p - 1}{2}\rceil$ are redundant. Therefore, we only retain the components $m^f_{i,k}$ corresponding to the positive frequencies which corresponds to elements in the range of $0$ to $\lfloor\frac{p}{2}\rfloor$. This allows for a more efficient representation of the frequency message, reducing redundancy and facilitating the subsequent resizing step.

\begin{figure*}[t]
\centering
\subfigure[Original messages $m^o$]{
\begin{minipage}[t]{0.4\linewidth}
\centering
\includegraphics[width=\linewidth]{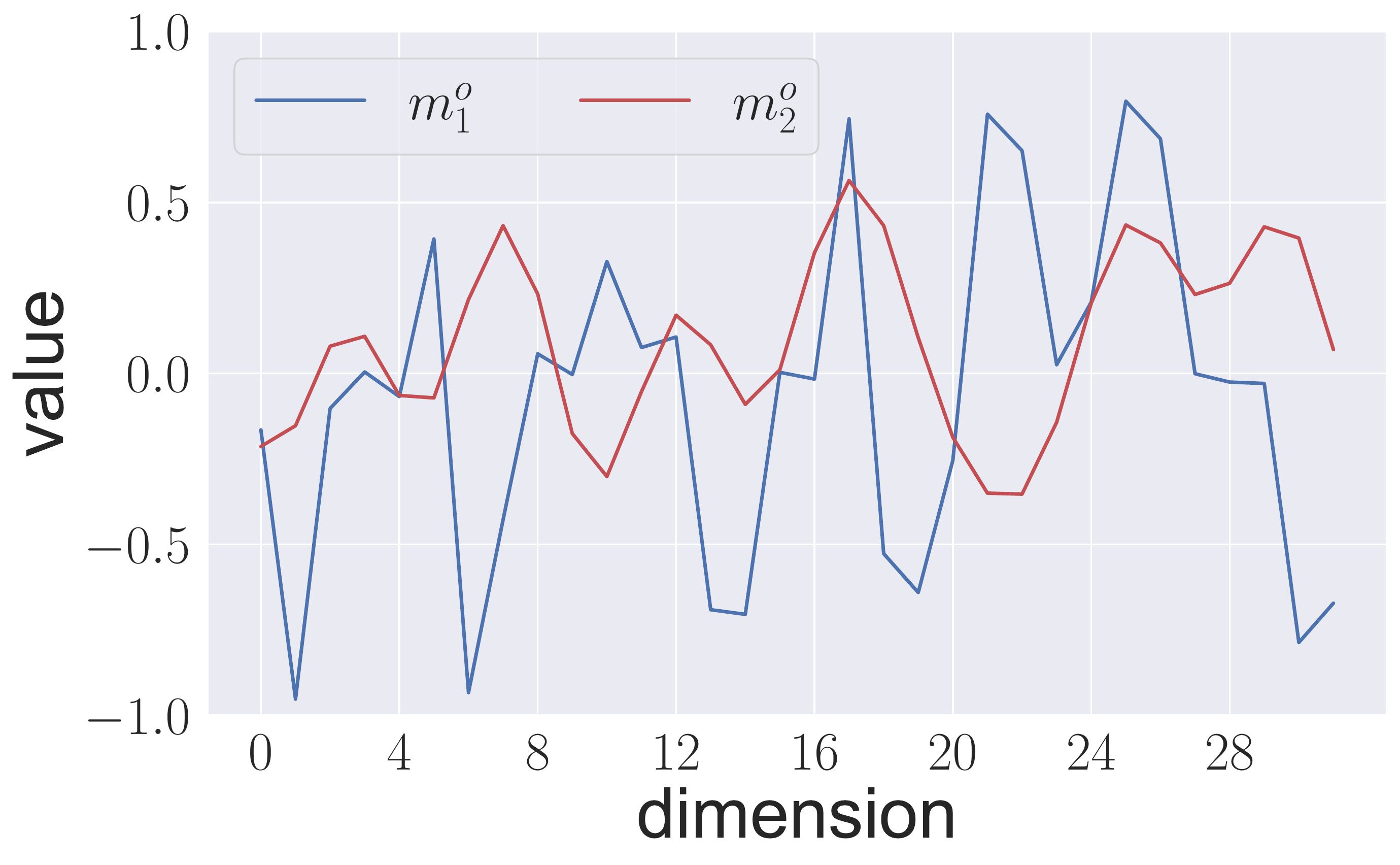}
\label{h1}
\end{minipage}
}
\subfigure[Frequency messages $m^f$]{
\begin{minipage}[t]{0.4\linewidth}
\centering
\includegraphics[width=\linewidth]{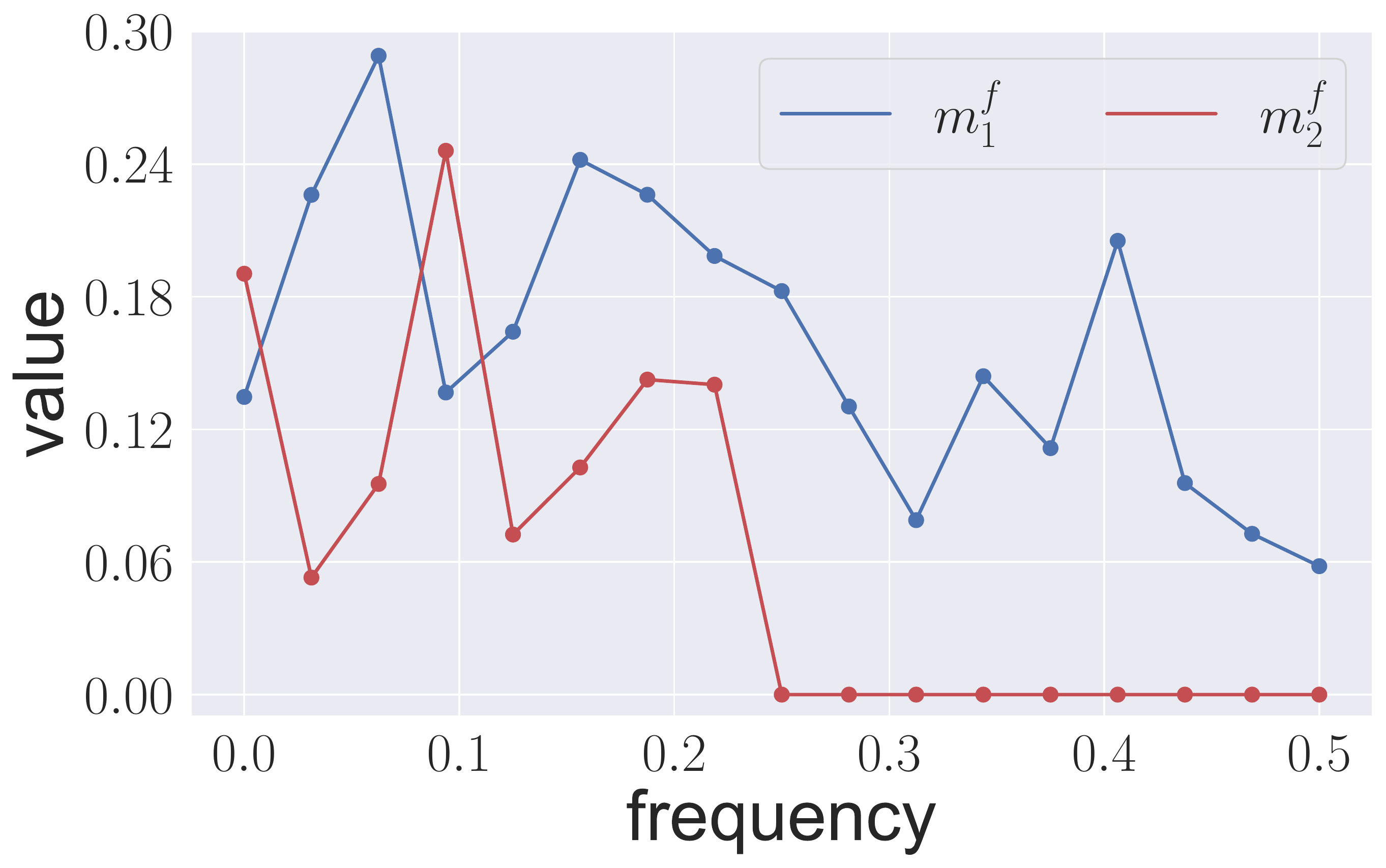}
\label{transformed_h1}
\end{minipage}%
}
\caption{Fourier Transform analysis of messages}
\label{example_clipping}
\end{figure*}

\smallparagraph{Clipping.} 
In order to match the allocated bandwidth $b_i$, the frequency message $m^f_i$ still needs to undergo clipping. Although clipping results in the removal of some information from the message, the DFT yields an inherent ordering of the importance of each element in the frequency domain. Since $m^f_i$ is a vector of complex numbers, transmitting each element requires $2$ bandwidth units — one for the real part and another for the complex part. Hence, we retain only the first $b_i / 2$ components of $m^f_i$, resulting in the clipped message $m^c_i$, which is sent to the other agents.

Figure \ref{example_clipping} illustrates two messages $m^o$, of size $p=32$, along with their frequency counterparts, $m^f$. These messages are obtained from trained policies. Notably, the frequency message $m^f_2$ exhibits an interesting property where the magnitudes after the 9th frequency are very close to zero. This demonstrates that by combining the DFT with clipping, the message can be effectively compressed, resulting in minimal loss of information in certain cases.

\subsection{Message interpretation and decision making}
\smallparagraph{Inverse Discrete Fourier Transform (IDFT)}~
is the mathematical operation that enables the reconstruction of the original message $m^r_i$ of agent $i$ from its clipped frequency version, $m^c_i$. Although the reconstruction may not be perfect, it retains the most significant components of the message. Importantly, the fidelity of the reconstruction improves as the allocated bandwidth increases. 

\smallparagraph{Decision making}~
The incoming messages received by agent $i$ are concatenated into $m_i$ by averaging them:
\begin{align}
    m_i = \frac{1}{N - 1} \sum_{j \neq i} m^r_i
\end{align}
The aggregated message $m_i$ is then incorporated into agent $i$'s decision-making process by conditioning its policy on the concatenation of its history $h_i$ and the aggregated message $m_i$, denoted as $\langle h_i, m_i \rangle$. Furthermore, the message $m_i$ is also included in the input of the LSTM in the next timestep, ensuring that agents memorize the important information contained in the received messages. This allows agents to make informed decisions based on both their individual history and the aggregated messages from other agents.

\subsection{Training}
The training of all the components in DSMS is conducted end-to-end using the MADDPG algorithm. In the DSMS framework, the critic network plays a crucial role, as it takes as input the weights $\bm{w}$ of the scheduler. This allows the critic to provide valuable feedback to both the scheduler and the agents, enabling them to learn and improve their communication strategies collaboratively. By training the entire DSMS system end-to-end, the components can jointly optimize their performance and adapt to the dynamics of the environment, resulting in more effective and efficient communication.

\begin{figure}[ht]
  \begin{minipage}[t]{0.5\linewidth}
    \raggedright
    \captionsetup{justification=raggedright,singlelinecheck=false}
    \captionof{table}{Results on Predator Prey}
    \begin{tabular}{ll}
    \hline
    Methods    & Collisions \\ \hline
    DSMS                          &  11.42                     \\ \hline
    MD-MADDPG                       & 1.97                       \\ \hline
    SchedNet                       & 0.97                     \\ \hline
    \label{table_pp}
    \end{tabular}
    \captionof{table}{Results on Cooperative Navigation}
    \begin{tabular}{lllll}
    \cline{1-3}
    Methods               & \begin{tabular}[c]{@{}l@{}}Avg.Dis\end{tabular} & Collisions(\%) &  &  \\ \cline{1-3}
    DSMS                  & 0.799                                              & 1.45          &  &  \\ \cline{1-3}
    MD-MADDPG             & 1.074                                              & 1.72          &  &  \\ \cline{1-3}
    SchedNet              & 1.581                                              & 1.16          &  &  \\ \cline{1-3}
    \label{table_cn}
\end{tabular}
  \end{minipage}%
    \hfill
  \begin{minipage}[t]{0.5\linewidth}
    \centering
    \subfigure[Predator Prey]{
    \begin{minipage}[t]{0.5\linewidth}
    \centering
    \includegraphics[width=3cm]{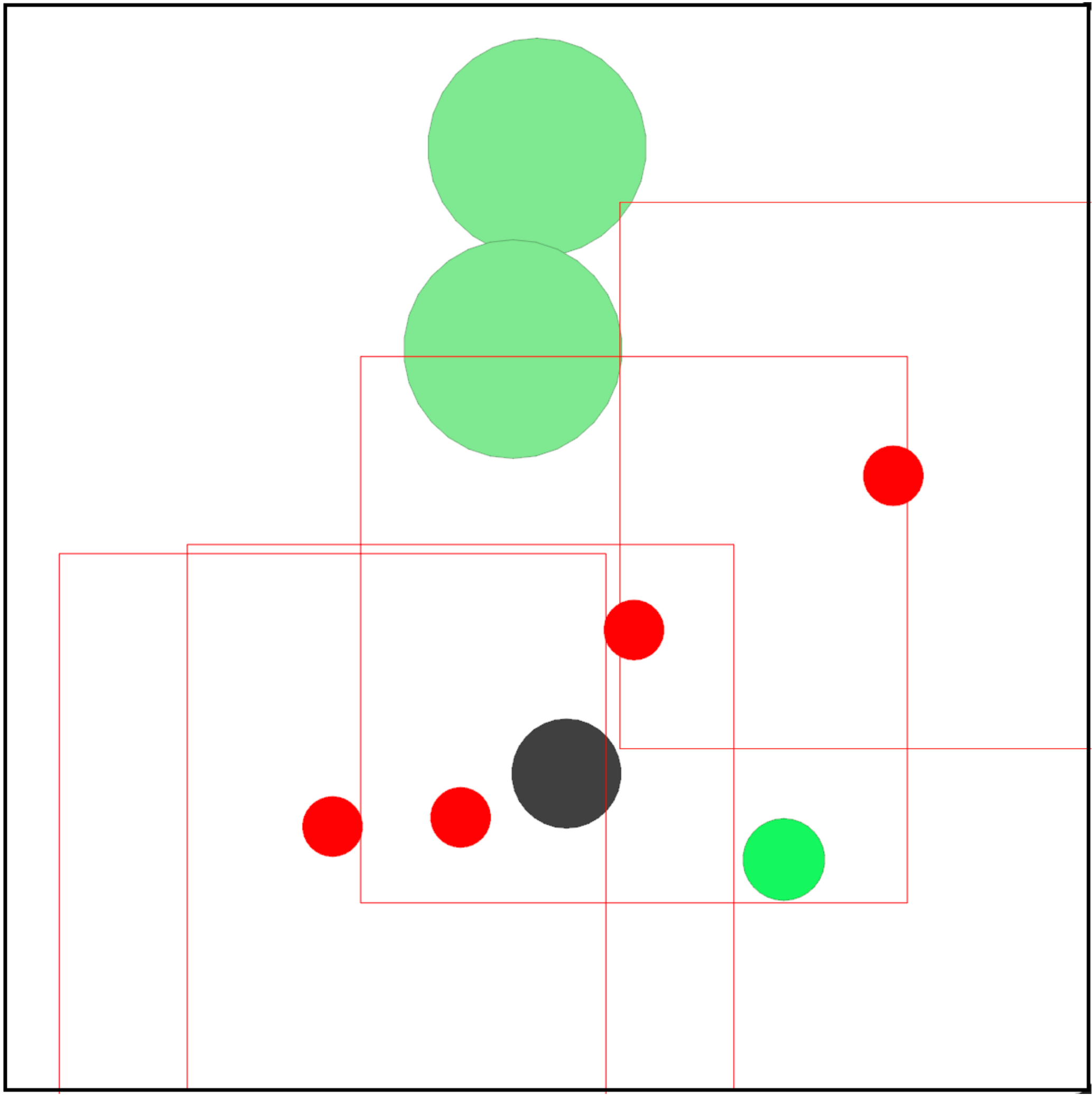}
    \label{pp_env}
    \end{minipage}%
    }%
    \subfigure[Cooperative Navigation]{
    \begin{minipage}[t]{0.5\linewidth}
    \centering
    \includegraphics[width=3cm]{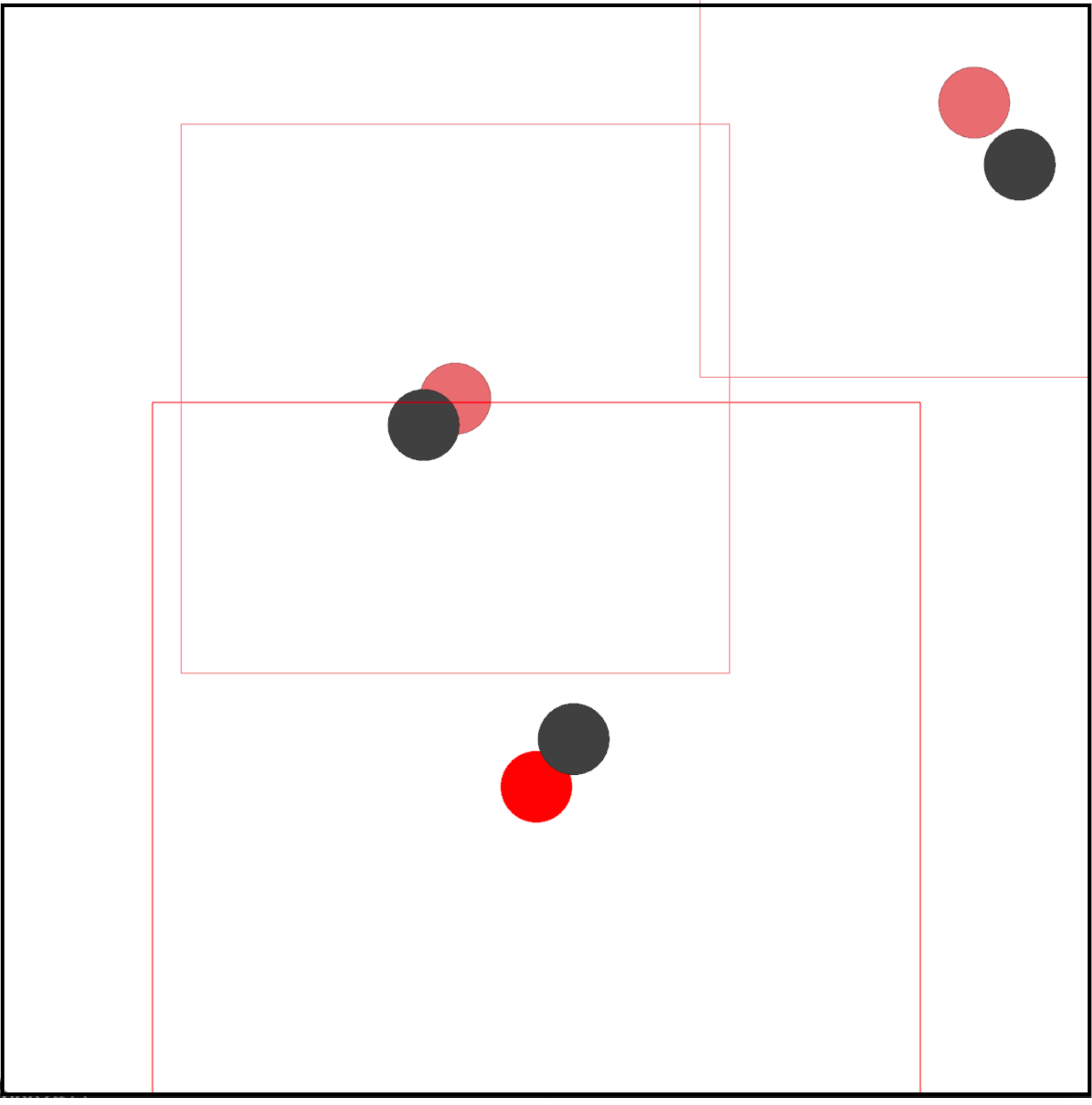}
    \label{cn_env}
    \end{minipage}%
    }%
    \caption{Environments}
    \label{image}
  \end{minipage}
\end{figure}
\section{Experiments}
\subsection{Environments}\label{environment}
We evaluate the performance of DSMS using two scenarios from the widely-used multi-agent particle environments \cite{lowe2017multi}: Predator Prey and Cooperative Navigation. These scenarios involve agents operating in a two-dimensional world with continuous state space. To add complexity and simulate real-world conditions, we modified these scenarios to impose partial observability.  

\smallparagraph{Predator Prey}~
In the Predator Prey scenario, our system consists of four predator agents (red) pursuing a single prey agent (green) as shown in Figure~\ref{pp_env}. The environment includes one landmark obstacle and two "forests" represented by large green balls, which can provide concealment for the agents. The prey agent has full observability and moves at the same velocity as the predators. It follows a fixed policy obtained by training it against DSMS predators with full observability.
For the predators, partial observability is introduced, where each agent's observation area is limited to a square centered on itself, with a length equal to half the size of the environment. This means that the maximum observation area is 25\% of the total environment when a predator is positioned at the center. Within their observation area, the predators can observe the relative positions and velocities of other agents.
The predators receive a shared reward, composed of positive components for every collision with the prey (+5) and a negative component based on the distance between each predator agent and the prey.

\smallparagraph{Cooperative Navigation}~
In the Cooperative Navigation scenario, our system consists of three agents (red) and three landmark objects (black) in the environment as shown in Figure~\ref{cn_env}. The objective is for the agents to cooperate with each other to cover all the landmarks without collisions.
Each agent has a partial field of observation, which allows them to perceive the relative positions of other agents and landmarks within their observation area. The "leader" agent, represented by a darker shade of red, has a larger observation area that covers up to 50\% of the environment when positioned at the center. The other two agents have observation areas that cover up to 25\% of the environment.
Individual rewards are assigned to each agent based on their distance to the closest landmark, penalties for collisions with other agents, and the distances between each landmark and the closest agent.

\subsection{Results}
\smallparagraph{Baselines}~ 
In our comparative analysis, we evaluated DSMS in comparison to two recent existing works: MD-MADDPG \cite{pesce2020improving} and SchedNet \cite{kim2019learning}, which serve as baselines for our study. These approaches specifically address the challenge of multi-agent communication over a single shared medium. MD-MADDPG introduces a shared memory space that agents can access sequentially, effectively reducing conflicts but granting all agents full access to the communication channel. On the other hand, SchedNet incorporates a scheduler that manages the shared communication medium and allocates limited bandwidth to agents, thereby determining which agents can utilize the channel for message transmission. By comparing DSMS with these baselines, we aim to assess its effectiveness and improvements in enhancing communication and coordination among agents.

\begin{figure}
\centering
\subfigure[Predator Prey]{
\begin{minipage}[t]{0.4\linewidth}
\centering
\includegraphics[width=\linewidth]{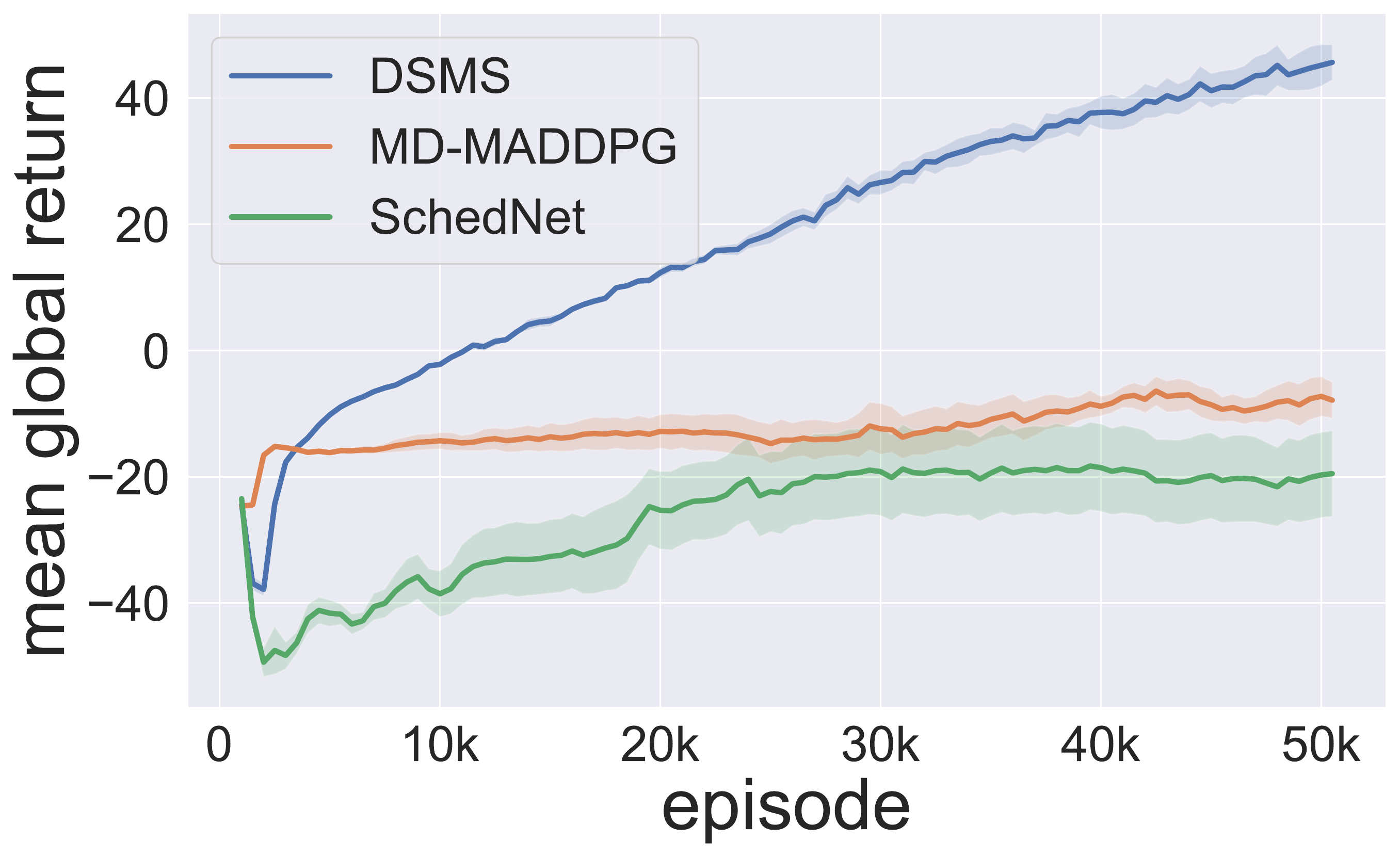}
\label{pp_baselines}
\end{minipage}%
}%
\subfigure[Cooperative Navigation]{
\begin{minipage}[t]{0.41\linewidth}
\centering
\includegraphics[width=\linewidth]{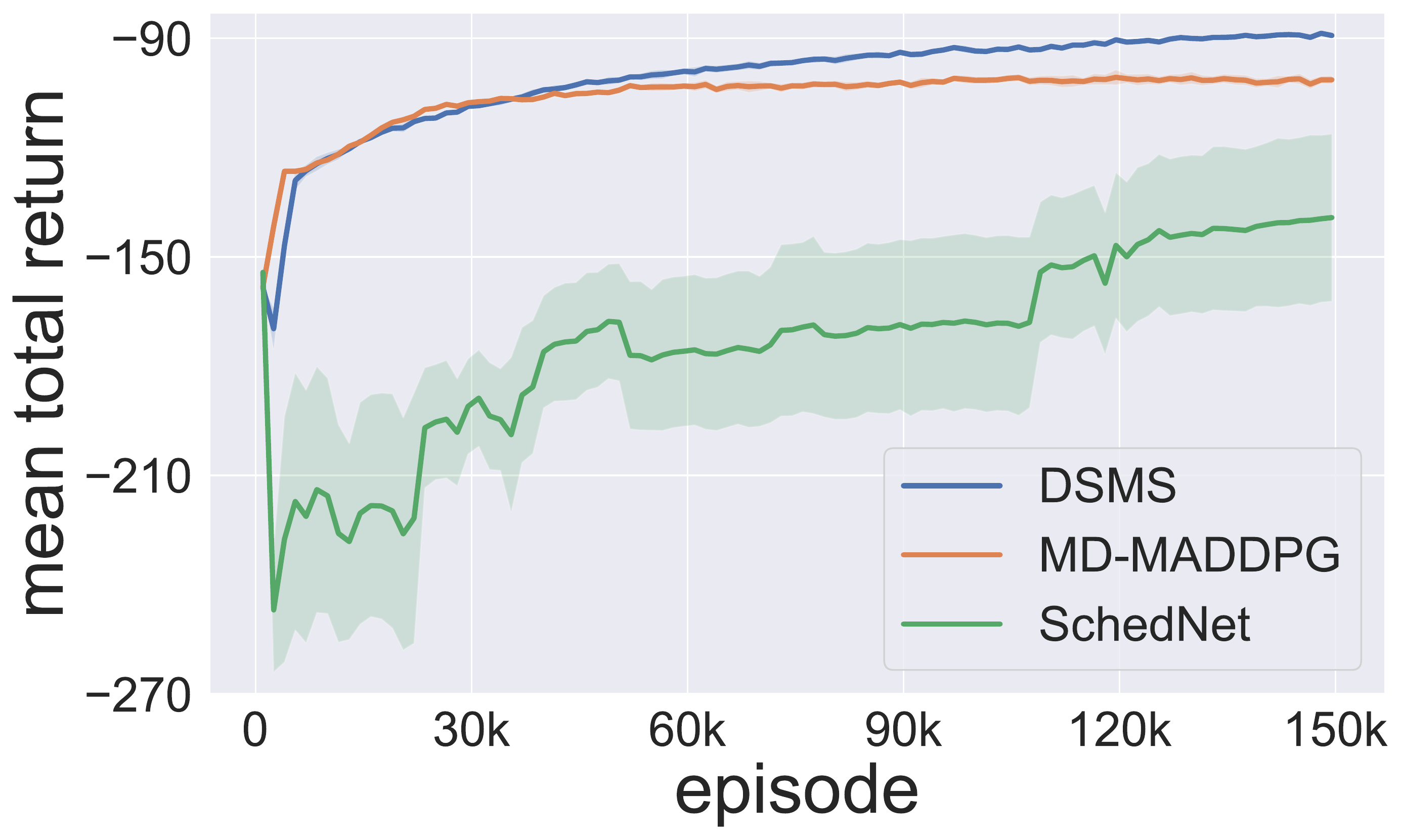}
\label{cn_baselines}
\end{minipage}%
}%
\caption{Evolution of the mean total return (sum of the agent's return) for Cooperative Navigation and mean global return (agent's receive the same reward) for Predator Prey during the training for DSMS, SchedNet and MD-MADDPG.}
\label{fig:results}
\end{figure}

In this subsection, we trained the models with a bandwidth of $B=64$, which allowed SchedNet to allocate two agents at each timestep. We conducted training for 50,000 episodes for the Predator-Prey scenario and 150,000 episodes for the Cooperative Navigation scenario. The results, averaged over $5$ different seeds, are presented in Figure \ref{fig:results}. 
For the Cooperative Navigation scenario, we measured the sum of the individual returns for each agent. This metric provides an indication of the overall performance of the team in completing the task. In the case of Predator-Prey, as the agents share the same reward, we plotted the global return.
Figure \ref{fig:results} illustrates the learning progress of DSMS compared to the baselines, MD-MADDPG and SchedNet. Tables \ref{table_pp} and \ref{table_cn} present different statistics of the final policies for the two scenarios. Those statistics were obtained by running 200 episodes for each seed (1,000 episodes in total).

\smallparagraph{Predator Prey}~ 
In the Predator-Prey scenario, DSMS exhibited faster learning compared to the other baselines as illustrated in Figure \ref{pp_baselines}. It achieved a final performance with an average total return of 40, while MD-MADDPG and SchedNet failed to obtain positive rewards, with respective scores of -10 and -20.
Table \ref{table_pp} provides insights into the final performance by presenting the frequency of capturing the prey during an episode. Both MD-MADDPG and SchedNet captured the prey less than 2.5 times per episode on average. In contrast, DSMS significantly outperformed them with an average capture rate of 11.42. This highlights the superior performance and effectiveness of DSMS in the Predator-Prey scenario.

\smallparagraph{Cooperative Navigation}~
In the Cooperative Navigation scenario, Figure \ref{cn_baselines} illustrates the learning curves of DSMS and the baselines. It is evident that DSMS and MD-MADDPG exhibit similar learning patterns in the initial episodes, but DSMS outperforms MD-MADDPG in terms of the final result. Notably, DSMS continues to improve even after 50,000 episodes, indicating its potential for further learning, while MD-MADDPG appears to have converged. 
In contrast, SchedNet faces difficulties in learning and falls behind the other algorithms in terms of performance. This suggests that the communication mechanism employed by DSMS is more effective and yields better results. Table \ref{table_cn} provides additional insights into the policies obtained after training. It reveals that the agents trained with DSMS approach the landmarks more closely. In contrast to our DSMS approach, SchedNet's agents keep some distance from the landmarks leading to less collision, but this cautious behavior results in a significant hindrance to its final performance. These results validate the efficiency of our communication method and highlight the promising potential of our fine-grained bytewise scheduling approach.

\subsection{Communication analysis}

\begin{figure}[t]
\centering
\subfigure[Bandwidth allocation]{
\begin{minipage}[t]{0.49\linewidth}
\centering
\includegraphics[width=\linewidth]{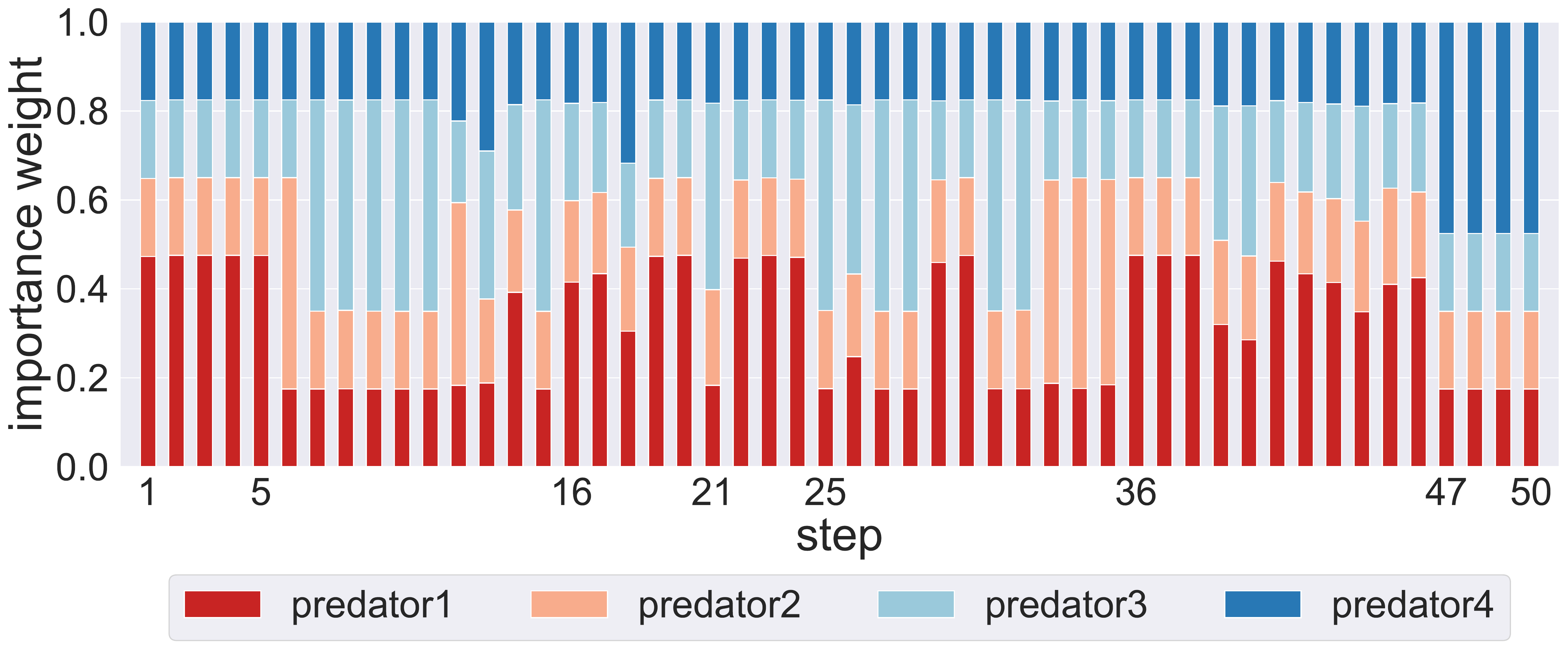}
\label{important_weight}
\end{minipage}%
}
\subfigure[Distances between predators and prey]{
\begin{minipage}[t]{0.49\linewidth}
\centering
\includegraphics[width=\linewidth]{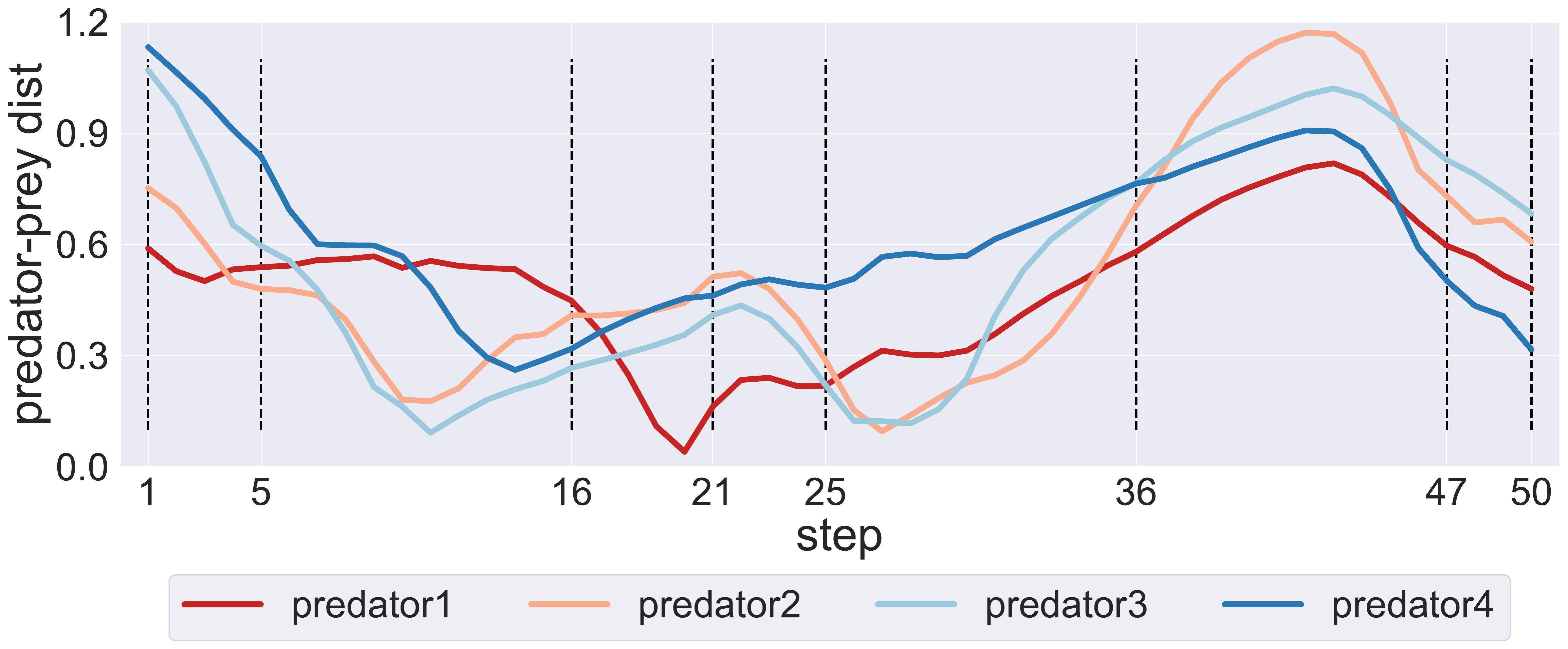}
\label{distance_prey}
\end{minipage}%
}
\caption{Communication analysis for Predator Prey}
\label{weitght_scheduling_pp}
\end{figure}

In this subsection, we analyze the division of the bandwidth between the agents.

\smallparagraph{Predator Prey}~
For this scenario, we compare the bandwidth repartition (Figure~\ref{important_weight}) with the distance between the agents and the prey (Figure~\ref{distance_prey}) during one episode of 50 steps. We separated the episode into seven blocks (1-4, 5-15, 16-20, 21-24, 25-35, 36-46, and 47-50). This separation reveals a correlation between the allocated bandwidth and the distance to the prey. In particular, the agent closer to the prey receives on average more bandwidth than the other, but it is not always the case. This shows that DSMS does indeed learn to attribute the bandwidth based on the importance of the message, and that the scheduling takes not only into account the current observation but also the novelty of the message which explains why agents more distant can receive more bandwidth.

\begin{wrapfigure}{r}{.4\linewidth}
\centering
\subfigure[Bandwidth allocation]{
\begin{minipage}[t]{\linewidth}
\centering
\includegraphics[width=\linewidth]{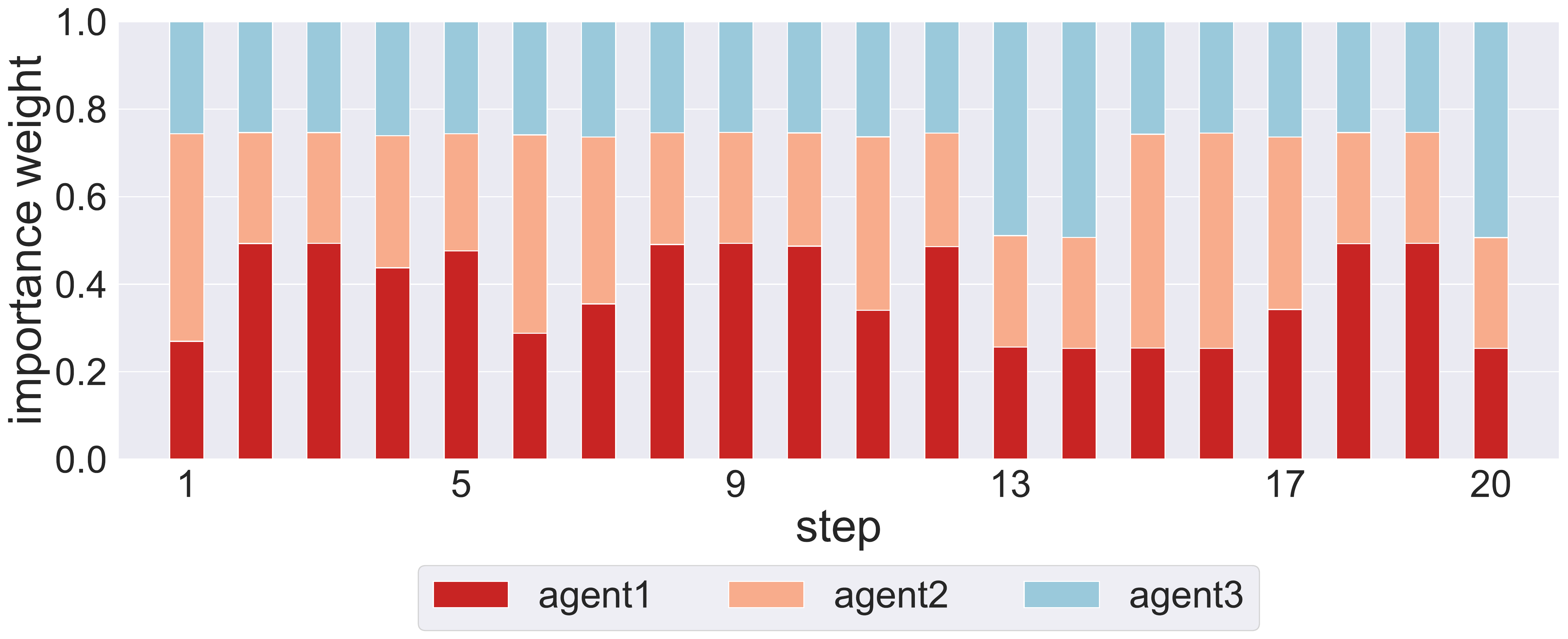}
\end{minipage}%
}
\caption{Communication analysis for Cooperative Navigation}
\label{weitght_scheduling_cn}
\end{wrapfigure}
\smallparagraph{Cooperative Navigation}~
For the Cooperative Navigation scenario, we focus on the allocation of bandwidth between the leader agent (Agent 1 in red) and the other two agents. In Figure~\ref{weitght_scheduling_cn}, we observe that the leader agent is allocated more bandwidth than the other two agents in 10 out of the 20 steps of the episode. This demonstrates that the DSMS scheduling mechanism is able to learn the importance of the leader's messages and allocate more bandwidth accordingly. By dynamically adjusting the bandwidth allocation based on the agent's role and observation window, DSMS effectively prioritizes the communication needs of the leader agent, leading to improved coordination and performance in cooperative navigation tasks.

\subsection{Ablations}

\begin{figure}
\centering
\subfigure[Predator Prey]{
\begin{minipage}[t]{0.45\linewidth}
\centering
\includegraphics[width=\linewidth]{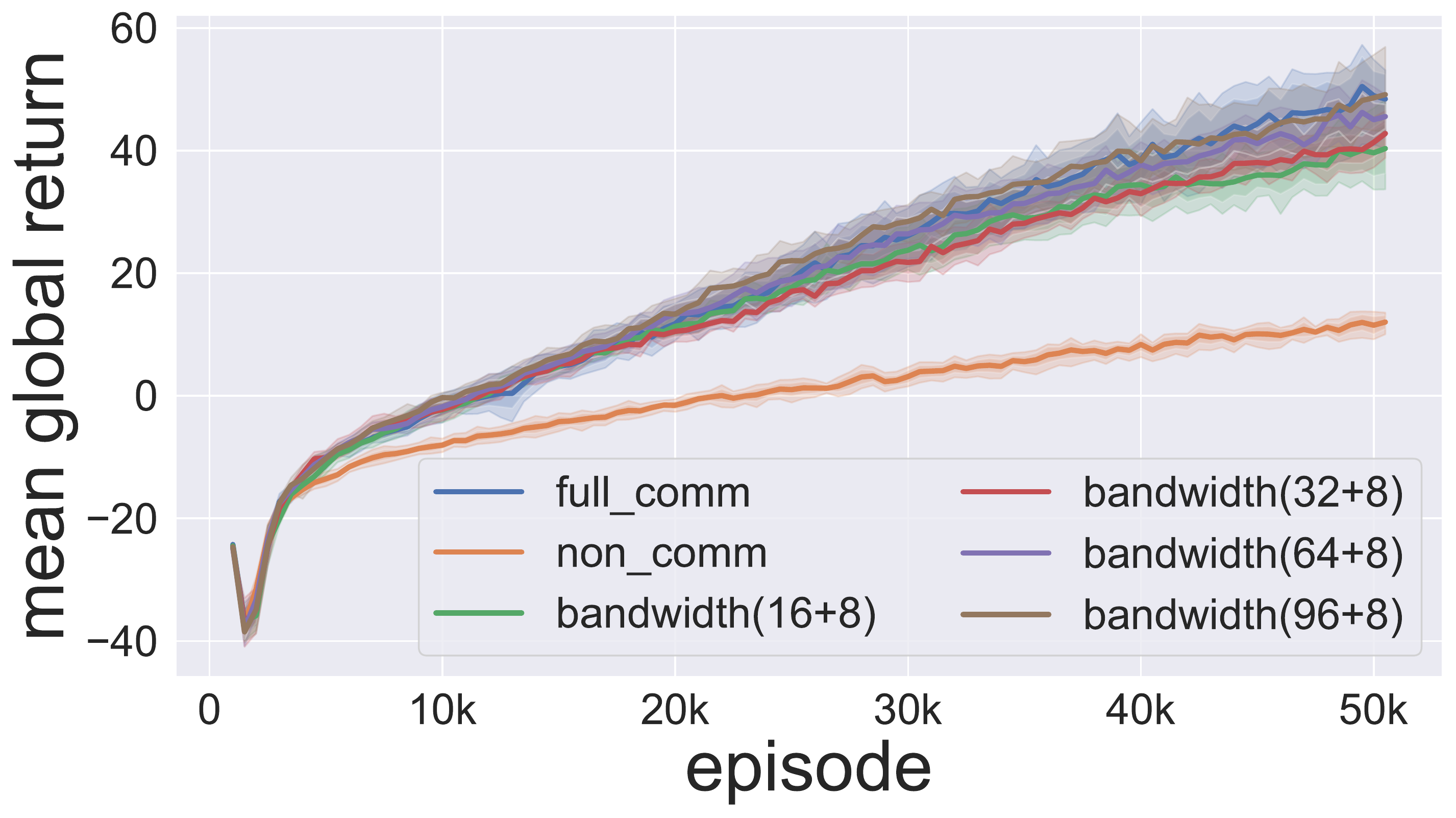}
\label{pp_ablation}
\end{minipage}%
}%
\subfigure[Cooperative Navigation]{
\begin{minipage}[t]{0.45\linewidth}
\centering
\includegraphics[width=\linewidth]{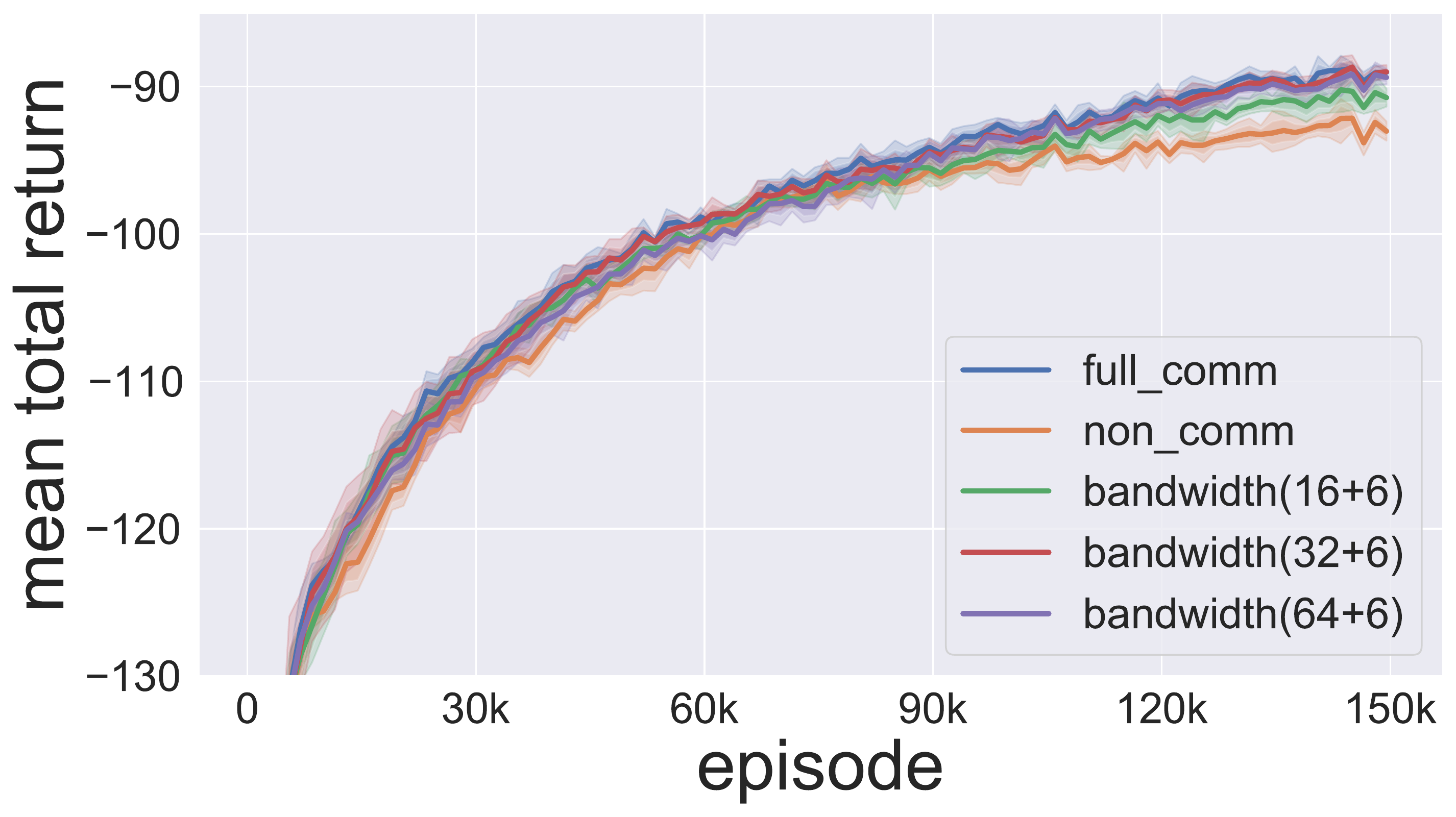}
\label{cn_ablation}
\end{minipage}%
}%
\caption{Evolution of the mean return during the training for DSMS with different bandwidth sizes including no communication and full communication.}
\label{fig:ablation}
\end{figure}
In addition to comparing DSMS with existing works, we conducted an ablative study to assess the impact of bandwidth on performance. We considered two extreme cases: no communication (bandwidth of 0) and full communication (bandwidth equal to the product of the number of agents with the size of a full message). Additionally, we tested bandwidth values of 24, 40, 72, and 104 for the Predator-Prey scenario, and 22, 38, and 68 for the Cooperative Navigation scenario. The Cooperative Navigation scenario features one agent less, hence the slightly lower bandwidth values.

The results of this ablative study, shown in Figure \ref{fig:ablation}, revealed several interesting findings. As expected, full communication achieved the best performance, while no communication yielded the worst performance. All non-zero bandwidth values demonstrated performance close to the full communication setting with a natural ordering based on the bandwidth size. This suggests that full communication is not necessary, and our approach performs well even with a small bandwidth allocation. These findings provide valuable insights into the optimal bandwidth requirements for effective multi-agent communication.

\section{Conclusions}

In this paper, we have presented a novel approach, Dynamic Sparse Message Scheduling, for addressing the challenges of communication in MARL systems. DSMS leverages Fourier Transform to dynamically resize messages, enabling a fine-grained scheduling mechanism to allocate bandwidth based on the importance of information.
We evaluated DSMS in two scenarios: Predator-Prey and Cooperative Navigation, with modifications to impose partial observability. In both scenarios, DSMS outperformed existing baselines, demonstrating faster learning and achieving higher final performance.
One important aspect of DSMS is its bandwidth allocation strategy, which dynamically adjusts the message sizes based on the importance of the information. Our analysis revealed that DSMS prioritizes agents with better observations or new information, leading to more efficient communication and coordination among agents.
Furthermore, an ablative study on different bandwidth sizes demonstrated that DSMS performs well even with reduced bandwidth. This highlights the robustness of our approach and its ability to adapt to varying communication constraints.
In conclusion, DSMS presents a promising solution to the communication challenges in MARL, offering improved coordination, faster learning, and effective bandwidth utilization. It paves the way for enhanced collaboration and communication among agents, enabling more sophisticated and efficient multi-agent systems in various domains.

Future work includes conducting additional experiments on different environments to further evaluate the performance of DSMS. We would also like to test DSMS with off-policy value-based methods such as VDN, QMIX or LAN \cite{singh2018learning, pmlr-v80-rashid18a, Avalos2022LocalLearningAAMAS}. Additionally, we plan to explore the replacement of the centralized scheduler with a decentralized approach, where agents reserve a subset of the bandwidth to share their next message utility. While this would improve decentralization, it would require agents to estimate their utility before taking an action and receiving the new observation, creating an offset in the communication protocol.

\subsection*{Acknowledgements}
This research was supported by funding from the Flemish Government under the ``Onderzoeksprogramma Artifici\"{e}le Intelligentie (AI) Vlaanderen'' program. This work was supported in part by the National Natural Science Foundation of China under Grants 62032018 and 61876151, in part by the Industry-University-Research Innovation Fund for the China Universities under Grant 2021ZYA09001, and by the China Scholarship Council.
R. Avalos is supported by the Research Foundation – Flanders (FWO), under grant number 11F5721N. 

\bibliographystyle{unsrt}
\bibliography{reference}
\end{document}